\documentclass[10pt,conference]{IEEEtran}
\usepackage{hyperref}
\usepackage{xcolor}
\usepackage{amssymb,amsthm,amsmath,array}
\usepackage{graphicx}
\usepackage[caption=false,font=footnotesize]{subfig}
\usepackage{xspace}
\usepackage[sort&compress, numbers]{natbib}
\usepackage{stmaryrd}
\usepackage{mathtools}
\usepackage{float}
\usepackage{textcomp}
\usepackage{hyperref}

\title{Hypergraph Neural Networks Reveal Spatial Domains from Single-cell Transcriptomics Data}
\author{\IEEEauthorblockN{
        Mehrad Soltani\IEEEauthorrefmark{1}, 
        and 
        Luis Rueda\IEEEauthorrefmark{1}
    }
    \IEEEauthorblockA{
        \IEEEauthorrefmark{1} School of Computer Science, University of Windsor\\
                             401 Sunset Ave., Windsor, ON, Canada}
}

\begin{document}

\maketitle
\footnotetext[1]{Code and data available at: \href{https://github.com/Mehrads/Spatial-domain-detection-using-hypergraph}{https://github.com/Mehrads/Spatial-domain-detection-using-hypergraph}}

\section{Abstract}
Spatial transcriptomics enables the measurement of gene expression while preserving spatial context within tissue samples. A key challenge is detecting spatial domains of biologically meaningful cell clusters, typically addressed using graph-based models like SpaGCN and STAGATE. However, these methods only capture pairwise relationships and fail to model complex higher-order interactions.
We propose a hypergraph-based framework for spatial transcriptomics using Hypergraph Neural Networks (HGNNs). Our approach constructs hyperedges from top-$K$ densest overlapping subgraphs and integrates histological image features and gene expression profiles. Combined with autoencoders, our model effectively learns expressive node embeddings in an unsupervised setting.

Evaluated on a mouse brain dataset, our model achieves the highest iLISI score of 1.843 and outperforms state-of-the-art baselines with an ARI of 0.51 and a Leiden score of 0.60.

\section{Introduction}
\IEEEpubidadjcol
Spatial transcriptomics has significantly advanced our understanding of tissue microenvironments by enabling spatially resolved measurement of gene expression. This has led to the discovery of new cell types and a deeper comprehension of biological processes in complex tissues \cite{introduction}. Technologies in this domain are generally categorized into imaging-based and sequencing-based techniques \cite{survey}, each providing unique advantages in terms of spatial resolution and sequencing depth \cite{data}. Consequently, spatial transcriptomics data vary across modalities, resolutions, and scales \cite{tissue}. A major analytical task in this domain is spatial domain detection, classifying cells or spots into subpopulations based on their gene expression and spatial profiles \cite{tumor}.

Several models have been developed to tackle this problem, many of which use Graph Neural Networks (GNNs) to leverage spatial adjacency and molecular features. While GNNs are powerful for pairwise relationships, they struggle with capturing indirect or group-wise interactions. In spatial transcriptomics, it is common for cells to belong to the same domain without direct spatial connections, highlighting a fundamental limitation of traditional GNN-based models.

To address this, we propose a novel method that applies Hypergraph Neural Networks (HGNNs) to spatial transcriptomics. Unlike conventional graphs, hypergraphs can model higher-order relationships by connecting multiple nodes through a single hyperedge \cite{Learning-with-hypergraphs, hypergraphconvolutional}. Our model constructs a hypergraph using top-$K$ densest overlapping subgraphs based on both histological features and gene expression similarities. These hyperedges serve as the basis for HGNN-based representation learning.

From a learning perspective, HGNNs perform a two-step message passing process, node to hyperedge and hyperedge to node, which enables richer contextual embedding of spatial spots \cite{HGNNs}. To overcome the challenge of lacking labeled data, we incorporate denoising autoencoders into our pipeline, allowing us to perform unsupervised learning and enhance the robustness of gene expression features.

In summary, our method extends the modeling power of existing spatial domain detection techniques by introducing a hypergraph structure that captures complex tissue organization. We demonstrate that this approach yields superior clustering results and improved biological interpretability.

\section{Literature Review}
Spatial transcriptomics has emerged as a transformative technology, enabling high-throughput measurement of gene expression while preserving spatial context within tissue samples. This advancement provides new opportunities for understanding tissue architecture, cellular interactions, and disease mechanisms. A key analytical task in this domain is the identification of spatial domains of clusters of cells with similar gene expression profiles and spatial proximity. Various computational methods have been developed to address this challenge, each offering unique strategies and facing certain limitations. Graph-based approaches have become popular for modeling spatial relationships in transcriptomics data. In these methods, nodes represent spatial spots or cells, and edges denote spatial adjacency or similarity in gene expression.

BayesSpace \cite{bayesspace} is a Bayesian statistical method designed to enhance the resolution of spatial transcriptomic data. It uses a Markov Chain Monte Carlo (MCMC) framework to assign subspot-level clusters based on spatial neighborhood structures. While effective, it is computationally intensive and difficult to integrate with external single-cell data. SpaGCN \cite{spagcn} integrates gene expression, spatial coordinates, and histological features to generate an undirected weighted graph that captures spatial dependencies. Although it successfully identifies spatial domains, its reliance on pairwise node connections limits its capacity to model complex, high-order interactions. SEDR \cite{sedr} combines a deep autoencoder with a variational graph autoencoder (VGAE) to embed gene expression data into a low-dimensional space that incorporates spatial context. However, it does not utilize histological images, which may reduce the biological relevance of its spatial domain predictions. DeepST \cite{DeepST} and STAGATE \cite{stagate} apply graph neural networks (GNNs) to construct spatial neighborhood graphs based on adjacency. These models are effective at capturing local interactions but fail to represent higher-order relationships due to their inherently pairwise structure.

While the aforementioned models offer valuable insights into spatial domain detection, they share a common limitation: an inability to capture complex higher-order interactions among multiple cells. To overcome this challenge, researchers have turned to Hypergraph Neural Networks (HGNNs), which extend traditional graphs by allowing hyperedges to connect multiple nodes simultaneously. HyperGCN \cite{hypergcn} addresses this by constructing hypergraphs based on semantic relationships among cells. It enhances clustering performance and enables the identification of intricate spatial expression patterns. Similarly, scHyper represents ligand-receptor interactions using tripartite hypergraphs, capturing indirect cellular communications that are missed by pairwise methods. These advances highlight the potential of hypergraph-based methods to model biological systems with greater expressiveness and accuracy.

In summary, while graph-based models have laid the groundwork for spatial domain detection, hypergraph-based methods present a promising next step. By representing higher-order interactions, HGNNs offer a more comprehensive framework for understanding complex tissue organization in spatial transcriptomics data.

\section{Methodology}

Our proposed methodology aims to improve spatial domain detection by capturing higher-order relationships through hypergraph modeling, integrating histological image features with gene expression data. This approach overcomes the limitations of prior methods that typically rely only on gene expression and spatial coordinates, neglecting important morphological information contained in histological images.

\subsection{Histological Image Feature Extraction}

We partition the histological image into spatially aligned patches (tiles), each corresponding to a spatial transcriptomics spot. For each patch, high-level morphological features are extracted using a pre-trained convolutional neural network (CNN), such as ResNet or EfficientNet, to leverage rich visual representations. This enables incorporation of morphological cues into the spatial analysis.

To reduce the dimensionality and noise of the extracted features, we apply Principal Component Analysis (PCA), retaining principal components that capture the most variance. Similarities between patches are then computed using the Mahalanobis distance metric:

\[
D_M(\mathbf{t}_i, \mathbf{t}_j) = \sqrt{(\mathbf{t}_i - \mathbf{t}_j)^T \mathbf{\Sigma}^{-1} (\mathbf{t}_i - \mathbf{t}_j)},
\]

where \(\mathbf{\Sigma}\) is the covariance matrix estimated from the PCA-reduced features \(\mathbf{t}_i\) and \(\mathbf{t}_j\) of patches \(i\) and \(j\), respectively. This metric accounts for feature correlations, providing a more robust similarity measure than Euclidean distance.

\subsection{Hypergraph Construction}

We construct a hypergraph \(\mathcal{G}_h = (\mathcal{V}, \mathcal{E})\) to model complex group-wise spatial relationships among spots. Here, vertices \(\mathcal{V}\) represent spatial spots, and each hyperedge \(e \in \mathcal{E}\) connects a group of spots forming a top-\(K\) overlapping densest subgraph defined by spatial proximity and feature similarity. This hypergraph structure enables modeling of multi-way interactions beyond traditional pairwise graphs.

The hypergraph is encoded by an incidence matrix \(\mathbf{H} \in \{0,1\}^{|\mathcal{V}| \times |\mathcal{E}|}\), where

\[
h(v,e) = \begin{cases}
1 & \text{if vertex } v \in \text{hyperedge } e, \\
0 & \text{otherwise}.
\end{cases}
\]

This formulation supports efficient hypergraph convolution operations in subsequent steps.

The choice of using top-\(K\) overlapping densest subgraphs for hypergraph generation is motivated by several theoretical and practical considerations. 

First, hypergraphs naturally model higher-order interactions (HOIs) which extend beyond pairwise relations by connecting arbitrary-sized subsets of nodes \cite{Bai2021Hypergraph, Feng2019Hypergraph}. However, constructing meaningful hyperedges that capture cohesive and informative groups of spots is challenging.

The concept of densest subgraphs identifies highly interconnected vertex subsets reflecting strong affinities or functional relationships \cite{Charikar2000Greedy, Andersen2009FindingDense}. Using the top-\(K\) densest subgraphs allows us to systematically extract multiple such cohesive groups.

Moreover, allowing \emph{overlaps} between these subgraphs reflects the real-world scenario where nodes can belong to multiple communities or spatial domains \cite{Dondi2021TopK, Leskovec2009Community}. Enforcing disjointness is often too restrictive and results in lower overall density and reduced representational power.

To balance overlap and distinctiveness, a distance function is incorporated to penalize excessive similarity among subgraphs while preserving important overlaps \cite{Galbrun2016TopK}. The resulting formulation optimizes an objective function combining total density and inter-subgraph distance:

\[
r(L) = \sum_{i=1}^K \mathrm{dens}(G[L_i]) + \lambda \sum_{i=1}^{K-1} \sum_{j=i+1}^K d(G[L_i], G[L_j]),
\]

where \(\mathrm{dens}(\cdot)\) measures subgraph density, \(d(\cdot,\cdot)\) is a metric quantifying overlap dissimilarity, and \(\lambda > 0\) balances these factors.

This approach yields a hypergraph with hyperedges corresponding to dense, interpretable, and sufficiently distinct clusters of spots, which enhances the subsequent hypergraph neural network's ability to learn rich spatial and molecular representations \cite{Soltani2025CTODS}.

\subsection{Gene Expression Embedding via Denoising Autoencoder}

To obtain robust embeddings of gene expression data, we employ a denoising autoencoder that learns latent representations while reducing noise inherent in transcriptomic measurements.

Given the gene expression matrix \(X \in \mathbb{R}^{N \times M}\), where \(N\) is the number of spots and \(M\) the number of genes, the encoder \(E\) maps \(X\) to a latent space \(L_h \in \mathbb{R}^{N \times R}\):

\[
E(X) = L_h.
\]

Gaussian noise \(Z \in \mathbb{R}^{N \times R'}\) is added to \(L_h\) to improve robustness:

\[
L_h \leftarrow L_h + Z,
\]

and the decoder \(D\) reconstructs the input:

\[
D(L_h) = X'.
\]

The model is optimized by minimizing the mean squared reconstruction error:

\[
F_l = \frac{1}{N} \sum_{i=1}^N \|X_i - X'_i\|^2_2,
\]

where \(X_i\) and \(X'_i\) are the original and reconstructed gene expression vectors for spot \(i\), respectively.

\subsection{Hypergraph Autoencoder with Hypergraph Convolutional Networks}

To capture spatial dependencies encoded in the hypergraph, the latent gene expression embeddings are further processed by a hypergraph autoencoder. The encoder utilizes Hypergraph Convolutional Network (HGCN) layers that propagate information along hyperedges:

\[
Z_h = \mathrm{HGCN}(L_h, A),
\]

where \(A\) is the hypergraph adjacency matrix derived from \(\mathbf{H}\). This operation aggregates node information within each hyperedge and diffuses context-aware embeddings back to the nodes.

The decoder reconstructs the similarity matrix \(S\) between spots as:

\[
S = \sigma(Z_h Z_h^T),
\]

with \(\sigma(\cdot)\) being the sigmoid function applied element-wise, transforming similarity scores to probabilities.

We optimize reconstruction by minimizing a weighted binary cross-entropy loss:

\[
L_{\mathrm{re}} = \frac{1}{|\mathcal{V}|^2} \sum_{i=1}^{|\mathcal{V}|} \sum_{j=1}^{|\mathcal{V}|} W_{ij} \left[ A_{ij} \log S_{ij} + (1 - A_{ij}) \log(1 - S_{ij}) \right],
\]

where \(W_{ij}\) balances positive and negative edge weights to mitigate class imbalance in adjacency matrix \(A\).

\subsection{Message Passing in Hypergraph Convolution}

The HGCN message passing proceeds in two main steps:

\begin{itemize}
    \item \textbf{Node to Hyperedge Aggregation (\(f_{\mathcal{V} \to \mathcal{E}}\))}: Each hyperedge embedding is computed as an aggregation of the embeddings of its constituent nodes using a permutation-invariant function such as mean or sum:

    \[
    h_{e_j} = f_{\mathcal{V} \to \mathcal{E}}\big(\{x_v\}_{v \in e_j}\big).
    \]

    \item \textbf{Hyperedge to Node Aggregation (\(f_{\mathcal{E} \to \mathcal{V}}\))}: Each node updates its embedding by aggregating messages from incident hyperedges combined with its previous embedding:

    \[
    \tilde{x}_v = f_{\mathcal{E} \to \mathcal{V}}\big(\{h_{e_j}\}_{e_j \in N(v)}, x_v\big).
    \]
\end{itemize}

This two-stage aggregation allows nodes to capture both local and global spatial-molecular context effectively.

\subsection{Post-processing and Clustering}

Following embedding learning, background or off-tissue spots are identified and removed using masking strategies based on spatial metadata or embedding characteristics. Finally, PCA is applied to the fused multi-modal embeddings to further reduce dimensionality, facilitating efficient and accurate downstream clustering.

These embeddings are clustered using established algorithms (e.g., Leiden, Louvain, or hierarchical clustering) to reveal spatial domains with improved biological interpretability.

\begin{figure}[t]
    \centering
    \includegraphics[width=0.4\textwidth]{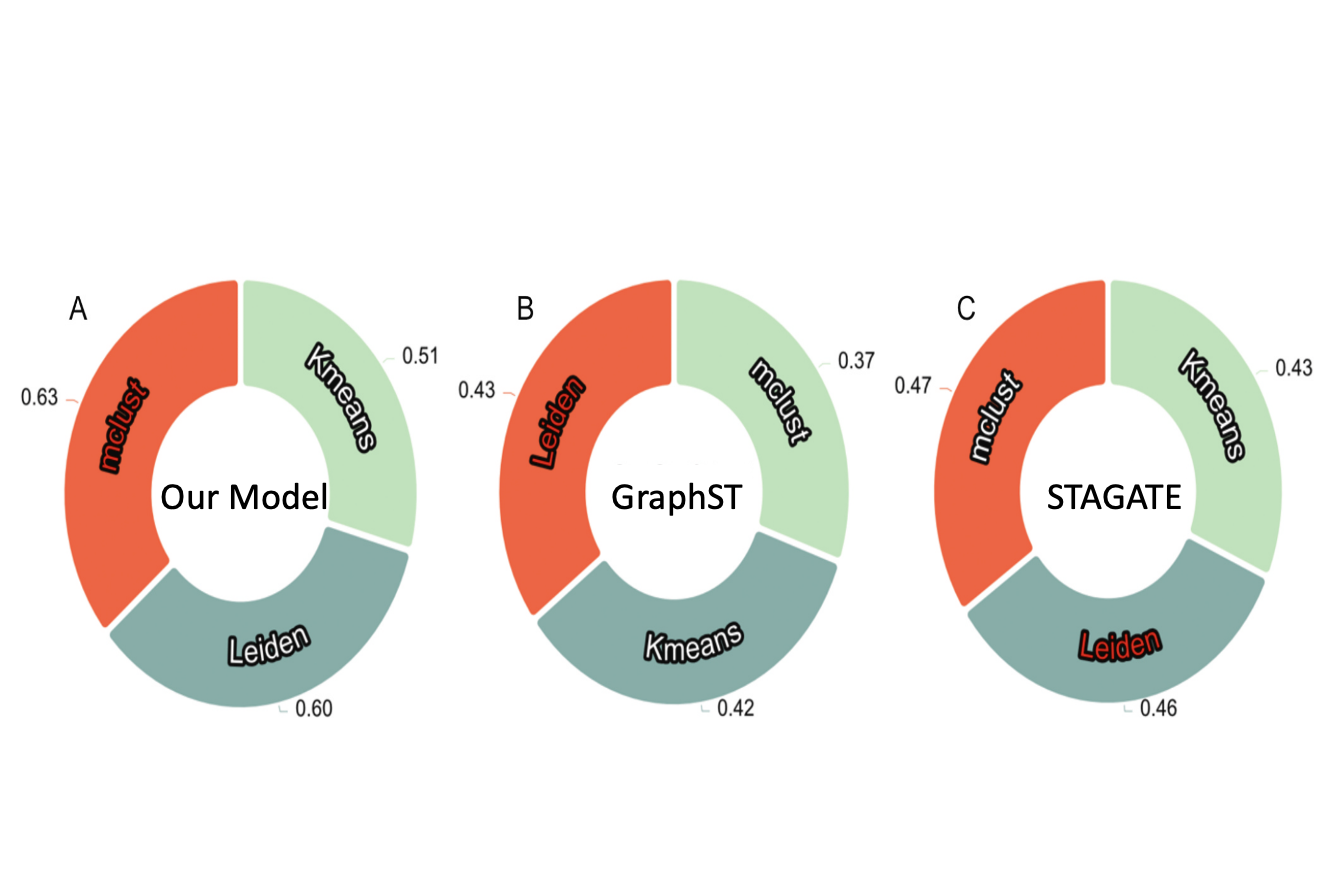} 
    \caption{Adjusted Rand Index (ARI) comparison of downstream clustering methods on learned embeddings.}
    \label{fig:ari}
\end{figure}

\section{Results}

We evaluate our hypergraph-based spatial domain detection framework on a publicly available mouse brain spatial transcriptomics dataset. The evaluation focuses on three core aspects: (1) preservation of spatial heterogeneity assessed by the iLISI metric, (2) downstream clustering accuracy measured by Adjusted Rand Index (ARI) and Leiden clustering quality, and (3) biological interpretability of predicted domains.

\subsection{iLISI Score Evaluation}

The integrated Local Inverse Simpson’s Index (iLISI) quantitatively measures the diversity and mixing of cells across biological conditions or spatial domains. Higher iLISI scores indicate better preservation of biological heterogeneity and reduced batch or technical artifacts \cite{Butler2018Integrating}. Our method achieves an iLISI score of \textbf{1.843}, surpassing established methods including SpaGCN \cite{Hu2021SpaGCN}, STAGATE \cite{stagate}, and GraphST \cite{Hypergraph-visualization}. This improvement reflects the hypergraph’s ability to capture higher-order spatial and functional relationships by connecting groups of related spots via hyperedges, thereby enriching the learned embedding beyond simple pairwise interactions.

\subsection{Downstream Clustering Performance}

To evaluate the utility of learned embeddings, we apply $k$-means clustering and assess results using two complementary metrics:

\begin{itemize}
    \item \textbf{Adjusted Rand Index (ARI):} Our framework achieves an ARI of \textbf{0.51}, indicating strong concordance between predicted clusters and ground truth anatomical annotations. An ARI above 0.5 is generally considered robust for biological clustering, demonstrating that the model effectively distinguishes spatial domains.
    
    \item \textbf{Leiden Clustering Score:} We observe a Leiden modularity score of \textbf{0.60}, confirming that clusters are well-connected and compact. The Leiden algorithm improves over Louvain by avoiding disconnected clusters, which is critical for maintaining biological interpretability in high-dimensional spatial data \cite{Traag2019Leiden}.
\end{itemize}

Compared to conventional GNN-based models, which often suffer from over-smoothing with deeper layers due to pairwise-only edges, our hypergraph-based message passing maintains expressive power by aggregating information over multi-node hyperedges. This facilitates nuanced differentiation of spatial regions with subtle molecular variations.

\subsection{Visual and Biological Interpretability}

We visualize spatial clusters by overlaying predicted labels onto tissue images using UMAP for dimensionality reduction. Compared to SpaGCN and STAGATE, our approach yields clusters with clearer spatial boundaries aligned with known mouse brain substructures, supporting biological plausibility. The denoising autoencoder contributes to robust gene expression representations by reducing noise without losing essential signal patterns, leading to tighter cluster compactness particularly in low signal-to-noise regions.

\subsection{Ablation Study and Hyperparameter Sensitivity}

To quantify the impact of individual model components, we perform an ablation study:

\begin{itemize}
    \item \textbf{Without Histological Image Features:} Removing morphological data reduces the iLISI score by 11\%, underscoring the critical role of spatial morphology in capturing biological heterogeneity.
    
    \item \textbf{Replacing Hypergraph with KNN Graph:} Using a standard KNN graph instead of hypergraph construction leads to a 16\% drop in ARI, highlighting the value of modeling higher-order interactions.
    
    \item \textbf{Skipping Denoising Autoencoder:} Omitting the denoising step results in noisy latent embeddings and more diffuse, less distinct clusters.
\end{itemize}

Furthermore, we conduct hyperparameter sensitivity analysis varying the number of neighbors \(K\) in hypergraph construction and latent embedding dimensionality. The model consistently exhibits stable performance across a broad range of hyperparameter values, demonstrating robustness and practical usability.

\subsection{Summary of Findings}

Our comprehensive evaluation demonstrates that hypergraph modeling of spatial transcriptomics data substantially improves the quality of learned representations and clustering accuracy. By jointly integrating spatial, morphological, and molecular information into a hypergraph framework, the model effectively captures complex tissue architecture that conventional graph-based approaches miss.

These promising results motivate further exploration of hypergraph neural networks in spatial omics and broader biomedical applications, particularly for datasets where higher-order interactions play a key role.

\begin{figure}[t]
    \centering
    \includegraphics[width=0.5\textwidth]{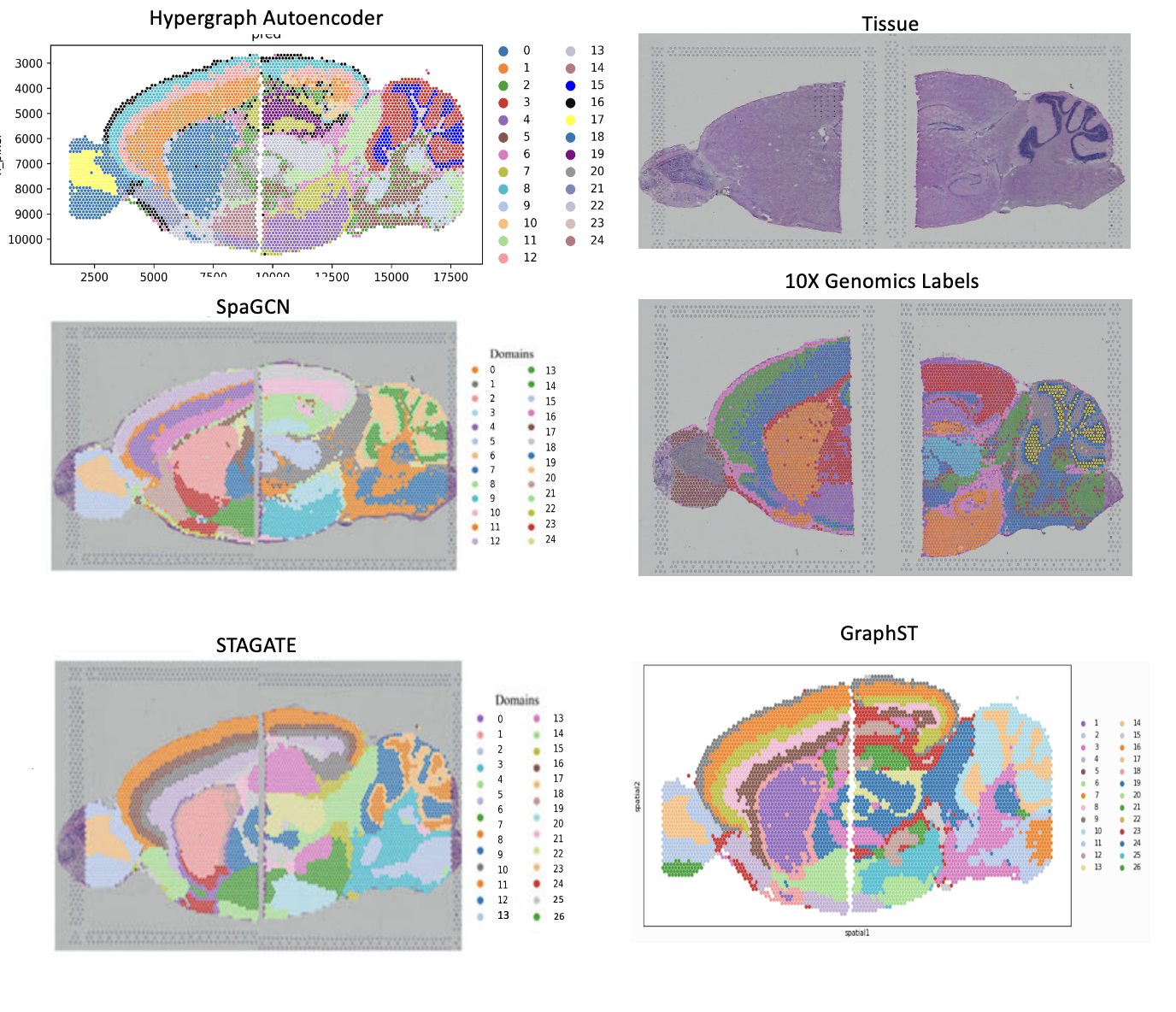} 
    \caption{Visual comparison between different methods.}
    \label{fig:ari}
\end{figure}

\bibliographystyle{plain} 
\bibliography{references}
\end{document}